\pdfoutput=1

\documentclass[conference]{IEEEtran}
\IEEEoverridecommandlockouts

\usepackage[utf8]{inputenc}
\usepackage{xcolor}
\usepackage{soul}
\usepackage[caption=false,font=footnotesize]{subfig}
\usepackage[ruled]{algorithm2e}
\usepackage{hyperref}
\usepackage{mathtools}
\usepackage[nolist]{acronym}
\usepackage[numbers]{natbib}
\usepackage{tabulary}
\usepackage{hhline}
\usepackage{graphicx}
\usepackage{fixltx2e}
\usepackage{array}
\usepackage{multirow}
\usepackage{numprint}
\npthousandsep{,}
\usepackage{makecell}
\usepackage[shortlabels]{enumitem}
\newcommand{\subparagraph}{}
\usepackage{titlesec}
\titlespacing*{\section}
{0pt}{1ex minus .5ex}{1ex minus 0.5ex}
\titlespacing*{\subsection}
{0pt}{1ex minus 1ex}{1ex plus .2ex minus 1ex}

\begin{document}
\title{\vspace{0em}Semantic Topic Analysis of Traffic Camera Images \vspace{-1.75em}
\thanks{This work was performed under the financial assistance award PSIAP3774 from U.S. Dept. of Commerce, National Institute of Standards and Technology}
\thanks{DISTRIBUTION STATEMENT A. Approved for public release. Distribution is unlimited. This material is based upon work supported by New Jersey Office of Homeland Security and Preparedness under Air Force Contract No. FA8702-15-D-0001. Any opinions, findings, conclusions or recommendations expressed in this material are those of the author(s) and do not necessarily reflect the views of the NJOHS. 
}
\thanks{We also acknowledge support from National Science Foundation grants CNS-1239054 and CNS-1453126, and FM IRG within the Singapore-MIT Alliance for Research and Technology.}
\thanks{\copyright 2018 IEEE. Personal use of this material is permitted. Permission from IEEE must be obtained for all other uses, in any current or future media, including reprinting/republishing this material for advertising or promotional purposes, creating new collective works, for resale or redistribution to servers or lists, or reuse of any copyrighted component of this work in other works.}
}
\author{\IEEEauthorblockN{Jeffrey Liu}
\IEEEauthorblockA{\textit{Civil and Environmental Engineering} \\
\textit{Massachusetts Institute of Technology}\\
Cambridge, MA 02139--4301 \\
jeffliu@mit.edu}\\\vspace{-4em}
\and
\IEEEauthorblockN{Andrew Weinert}
\IEEEauthorblockA{\textit{Lincoln Laboratory} \\
\textit{Massachusetts Institute of Technology}\\
Lexington, MA 02421--642\\
andrew.weinert@ll.mit.edu}\\\vspace{-4em}
\and
\IEEEauthorblockN{Saurabh Amin}
\IEEEauthorblockA{\textit{Civil and Environmental Engineering} \\
\textit{Massachusetts Institute of Technology}\\
Cambridge, MA 02139--4301 \\
amins@mit.edu}\\\vspace{-4em}
}

\maketitle
\begin{abstract}
Traffic cameras are commonly deployed monitoring components in road infrastructure networks, providing operators visual information about conditions at critical points in the network. However, human observers are often limited in their ability to process simultaneous information sources. Recent advancements in computer vision, driven by deep learning methods, have enabled general object recognition, unlocking opportunities for camera-based sensing beyond the existing human observer paradigm. In this paper, we present a \acl{NLP}-inspired approach, entitled \ac{BoLW}, for analyzing image data sets using exclusively textual labels. The \ac{BoLW} model represents the data in a conventional matrix form, enabling data compression and decomposition techniques, while preserving semantic interpretability. We apply the \acl{LDA} topic model to decompose the label data into a small number of semantic topics. To illustrate our approach, we use freeway camera images collected from the Boston area between December 2017–January 2018. We analyze the cameras’ sensitivity to weather events; identify temporal traffic patterns; and analyze the impact of infrequent events, such as the winter holidays and the “bomb cyclone” winter storm. This study demonstrates the flexibility of our approach, which allows us to analyze weather events and freeway traffic using only traffic camera image labels.
\end{abstract}

\newcommand{\camera}{c}
\newcommand{\cameraset}{\mathcal{C}}
\newcommand{\dimreduce}{g}
\newcommand{\imageidx}{i}
\newcommand{\image}{I}
\newcommand{\imageset}{\mathcal{I}}
\newcommand{\corpus}{\imageset}
\newcommand{\labelword}{\lambda}
\newcommand{\termidx}{j}
\newcommand{\labelvec}{\ell}
\newcommand{\labelmat}{\Lambda}
\newcommand{\labelset}{\mathcal{L}}
\newcommand{\countimages}{n}
\newcommand{\numimages}{N}
\newcommand{\numlabels}{M}
\newcommand{\numclusters}{\numtopics}
\newcommand{\timestamp}{\tau}
\newcommand{\topic}{z}
\newcommand{\topicset}{\mathcal{Z}}
\newcommand{\topicidx}{k}
\newcommand{\numtopics}{K}
\newcommand{\docfreq}{f}
\newcommand{\weight}{w}
\newcommand{\topicwordparam}{\beta}
\newcommand{\topicworddist}{\phi}
\newcommand{\doctopicparam}{\alpha}
\newcommand{\doctopicdist}{\theta}

\begin{acronym}
\acro{API}{application programming interface}
\acro{CNN}{convolutional neural network}
\acro{GCV}[GCV]{Google Cloud Vision}
\acro{LDA}[LDA]{Latent Dirichlet Allocation}
\acro{LSI}[LSI]{Latent Semantic Indexing}
\acro{SVD}[SVD]{Singular Value Decomposition}
\acro{NLP}[NLP]{Natural Language Processing}
\acro{tf-idf}{Term Frequency-Inverse Document Frequency}
\acro{MassDOT}{Massachusetts Department of Transportation}
\acro{BoW}{Bag-of-Words}
\acroplural{BoWs}{Bags-of-Words}
\acro{BoLW}{Bag-of-Label-Words}
\acroplural{BoLW}[BoLWs]{Bags-of-Label-Words}
\acro{NOAA-GHCN}{NOAA Global Historical Climatology Network}
\acro{LS1}{Labeling Service 1}
\end{acronym}
\section{Introduction}
\label{sec:intro}
Monitoring the traffic state and driving conditions of a road infrastructure network is crucial for efficient operations. While embedded sensors such as loop detectors can give precise measurements of specific quantities, it is often impractical to install specific sensors for every quantity of interest. Thus, many transportation agencies have integrated cameras into their monitoring solutions. For example, one service provider, TrafficLand, operates over \numprint{18000} cameras across over $200$ cities in the United States \cite{TrafficLand2014}. Cameras offer several benefits: they are general purpose sensors that can detect multiple quantities in a region of space; they can be installed on existing physical and communications infrastructure; and human operators intuitively understand images. However, since humans are limited in their ability to process visual information from simultaneous information sources \cite{Marois2005}, it is unreliable to depend on human operators to constantly monitor all of the cameras. 

While computers excel at processing large quantities of structured data, images are unstructured, and have been historically difficult for machines to parse. Prior attempts to explicitly parameterize object detection for traffic applications faced difficulties dealing with the separation of foreground and background elements, the occlusion of objects, variability in lighting conditions, and computational costs \cite{Kastrinaki2003}\cite{Buch2011}. As a result, many of the successful applications of traffic cameras for automatic sensing have been limited to narrowly-defined problems, such as automatic license plate detection \cite{Buch2011}. The use of cameras in these applications more closely resemble single-purpose sensors, as they only detect a specific quantity---e.g. license plate numbers. Thus, outside of these narrow applications and direct observation by operators, the potential of cameras for automatic, general-purpose, sensing in traffic applications has not yet been fully realized.

Recent advancements in computational power and deep learning methods, particularly deep \acfp{CNN}, have driven remarkable progress in recognizing objects in images \cite{lecun2015deep}. These techniques infer rules for object detection based on large training data sets of labeled example images \cite{krizhevsky2012imagenet}. Once trained, predictions using \acp{CNN} are quick to perform and can be done even on mobile devices \cite{abadi2016tensorflow}. However, training neural networks is computationally expensive, and requires large quantities of labeled training data \cite{Livni2014a}. Thus, within the last few years, technology companies have begun to offer access to pre-trained image recognition algorithms as commercial services \cite{GoogleImageBeta}. These services allow developers to quickly obtain labels in plain English for any image, without needing to build and train their own image recognition system. 

Although there is significant attention currently focused on improving the performance of deep learning algorithms \cite{lecun2015deep}, our focus is on exploring new applications enabled by these technologies. To this end, we pose the motivating question: \emph{what operationally-relevant information about phenomena in a traffic network can be obtained using only the labels of traffic camera images}? We illustrate the question with a thought experiment: imagine a blindfolded observer who is continuously told verbally whether the camera currently shows an item from a finite list of recognized objects; no additional information is given about the number, location, nor appearance of the objects in the frame, only simply whether or not it appears in the image. What can the blindfolded observer infer about the network state? In particular: \emph{can the blindfolded observer distinguish between a new instance of a phenomenon versus the persistence of an existing one}, and \emph{can they discern differences in magnitude between phenomena}?

To address these questions, we present in Sec.~\ref{sec:methods} an approach to analyzing sets of images using only the textual labels from an image recognition service. By treating sets of image labels as ``documents" describing the images, we pose the problem as a \ac{NLP} problem of analyzing a corpus of texts. We introduce the \acf{BoLW} model, inspired by the popular \ac{BoW} model \cite{Manning2009a}, which represents the image content labels in a conventional matrix structure, allowing for data compression and dimension reduction operations. We demonstrate the application of \acf{LDA} \cite{Blei2003a}, a hierarchical Bayesian model of topics within text corpora, to decompose the label data into high-level semantic topics. We present a case study based on traffic camera images from the Boston area, described in Sec.~\ref{sec:data}. Sec.~\ref{sec:results} describes the case study results regarding the detection of weather events, and the impact on weekly traffic patterns caused by the winter holidays and the ``bomb cyclone" storm. Sec.~\ref{sec:discussion} concludes and presents directions for future work.


\section{Traffic Camera Data}
\label{sec:data}
\subsection{Cameras and Images}
\begin{figure}[!htb]
\centering
\subfloat[MassDOT camera IDs and names]{\footnotesize
\begin{tabular}{|c|l|}
    \hline
        \textbf{ID} & \textbf{Name}\\ \hline
        1106--1 & I-93-ME-Boston-@ exit to HOV-E \\ \hline
        1137--1 & I-93-NB-Charlestown-@ Zakim South Twr\\ \hline
        1296--1 & Ramp I-EB-TNL-ramp end 93N x20 c\\ \hline
        1413--1 & Ramp K-NB-Boston-93N x20 b  \\ \hline
        1500--1 & Ramp K-NB-Boston-93N x20 a  \\ \hline
        1508--1 & Ramp CC-EB-Boston-90E x24C to 93S e\\ \hline
        1600--1 & Road OHWY-SB-Boston-Leverett Circle\\ \hline
    \end{tabular}}\label{tab:camera_ids}
\subfloat[Camera locations and sample images]{\includegraphics[width=0.8\linewidth]{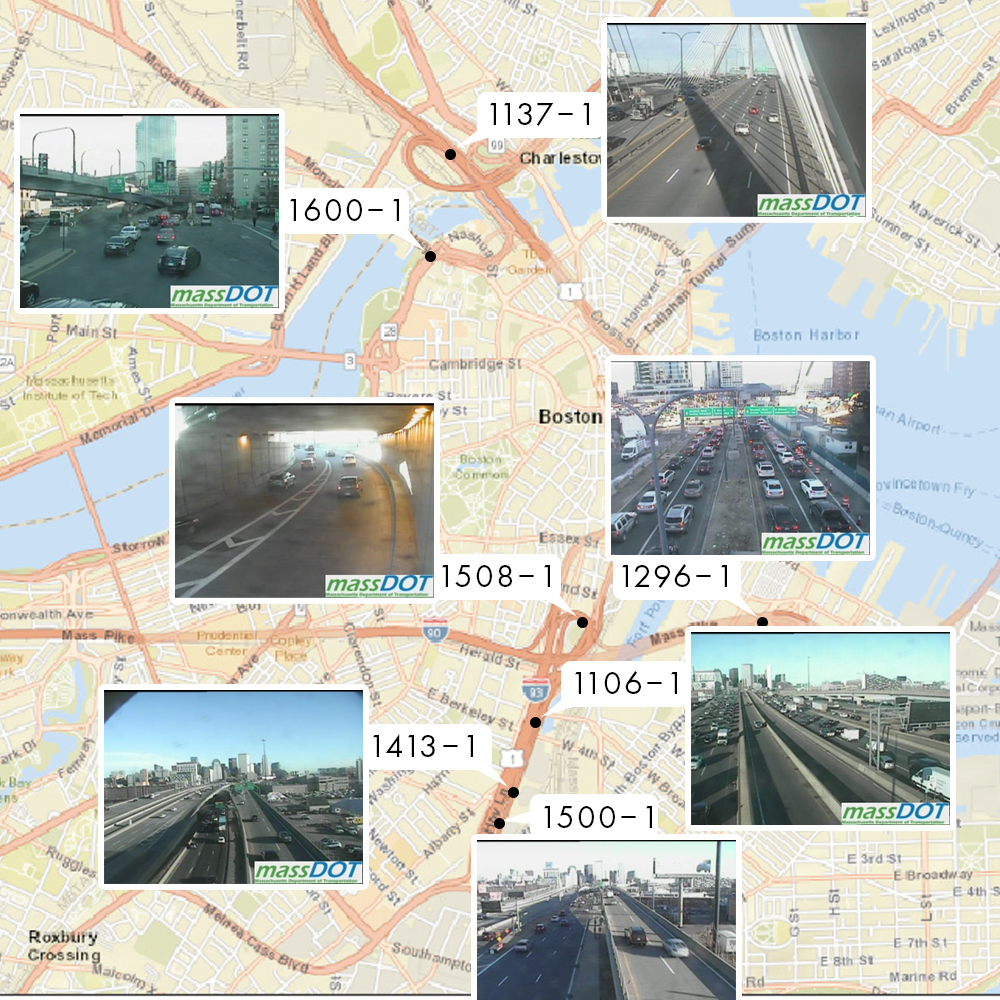}%
\label{fig:locations}}
\caption{Camera details. We selected a diverse set of cameras which depicted several different network locations and components, including a bridge (1137--1), underpass (1508--1), intersection (1600--1), HOV lane (1106--1), median (1296--1), and open freeway (1413--1, 1500--1)}
\label{fig:cameras}
\end{figure}

We illustrate our approach using data from cameras in the Boston area. The data consists of images collected from seven freeway traffic cameras operated by the \ac{MassDOT}. We regularly scraped the public Mass511 Traveler Information Service website \cite{Mass511Cams} between December 17, 2017--January 31, 2018 to build a data set of \numprint{189498} images, each with a resolution of $320\times240$ pixels, and an average sampling period of 3 minutes for each camera. Notable events during the collection period include the winter holidays and the ``bomb cyclone" East Coast blizzard. Details for each camera, including their \ac{MassDOT}-assigned identification number and name, locations, and sample images, are provided in Figure~\ref{fig:cameras}. 

\subsection{Labels}
\label{sub:labels}
We obtained labels for the traffic camera images using the \ac{GCV} image recognition platform\footnote{Note that the approach in this paper is not specific to the \ac{GCV} services, and any image labeling solution can be used in its place.}; in particular, we used the ``label detection" service---referred to as \ac{LS1}---and the ``web entity detection" service (resp. LS2). The label detection annotation service (LS1) provides annotations for ``broad sets of categories within an image, ranging from modes of transportation to animals," whereas the web detection service (LS2), provides more detailed information from the web, such as related websites and associated ``web entities" \cite{GoogleClientAPIDoc}---elements of the Google Knowledge Graph ontology representing real-world entities and concepts \cite{singhal2012introducing}. We use the names of these related web entities as a second source of image labels. We define the term ``vocabulary" of a service to represent the set of all labels that the particular service can return. In this paper, we consider a unified vocabulary, constructed from the disjoint union of the LS1 and LS2 vocabularies.

\begin{figure*}[htb]
\centering
\subfloat[Camera 1137--1, 2018-01-04 16:57:52 (UTC) ]{\raisebox{-0.7in}{\includegraphics[width=0.3\linewidth]{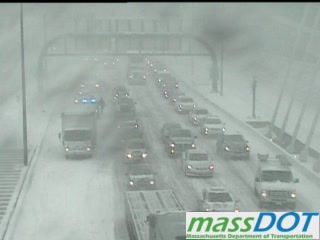}}%
\label{fig:annotation_example}}\hspace{0.5in}
\subfloat[Labels and scores for image \protect\subref{fig:annotation_example}]{
\footnotesize
\begin{tabular}{|m{0.15\linewidth}|m{0.05\linewidth}|}
\hline
    \textbf{LS1 label} & \textbf{Score}\\ \hline
        snow &	0.91\\\hline
    	infrastructure &	0.87\\\hline
    	mode of transport &	0.87\\\hline
    	lane &	0.84\\\hline
    	winter storm &	0.84\\\hline
    	road &	0.83\\\hline
    	transport &	0.82\\\hline
    	structure &	0.77\\\hline
    	phenomenon &	0.75\\\hline
    	blizzard &	0.7\\\hline
    	highway &	0.71\\\hline
    	freezing &	0.65\\\hline
    	automotive exterior &	0.57\\\hline
    	glass &	0.55\\\hline
\end{tabular}
\begin{tabular}{|m{0.1\linewidth}|m{0.05\linewidth}|}
    \hline
     \textbf{LS2 label} & \textbf{Score}\\ \hline
        Blizzard &	1.21\\\hline
        Lane   &	1.10\\\hline
        Car	 & 1.0\\\hline
        Transport &	1.03\\\hline
        Snow &	0.75\\\hline
        Highway	 & 0.75\\\hline
        Fog &	0.72\\\hline
        Glass &	0.67\\\hline
        Freezing &	0.62\\\hline
        \makecell[bl]{Massachusetts\\ Department of\\ Transportation} &	0.32\\ \hline
    \end{tabular}
\label{tab:labels}}
\caption{An image taken during the ``bomb cyclone" \protect\subref{fig:annotation_example} and the labels and scores returned by the labeling services \protect\subref{tab:labels} }
\label{fig:label_samples}
\end{figure*}
	
Labels and their respective scores for a sample image are given in Fig.~\ref{fig:label_samples}. Note that the labels from LS1 are reported by the service in lowercase, and those from LS2 are reported in Title Case; when necessary, we may also prepend the originating service to the label to further distinguish labels from the respective service, e.g. ``LS1: car" vs. ``LS2: Car". The score of each label corresponds to the confidence level reported by \ac{GCV}. The label scores from LS1 are reported by \ac{GCV} on a normalized scale, with a maximum value of 1 and truncated at 0.5, i.e. no labels are returned by the service whose scores is less than 0.5. The scores returned from LS2 are not normalized by \ac{GCV}, and the documentation warns that the scores should not be compared between labels nor images \cite{GoogleClientAPIDoc}. As such, we discard all score information and binarize the data by setting all nonzero scores to 1. We demonstrate that even with such an aggressively data processing approach, we are still able to clearly identify topics and phenomena.


In general, the LS2 vocabulary includes more specific terms than the LS1 vocabulary; for example, we observed the labels ``LS2: BMW," ``LS2: BMW 3 Series," and ``LS2: 2018 BMW 3 Series Sedan" in the vocabulary, whereas we found only the label ``LS1: bmw" in the LS1 vocabulary. However, the LS2 was also prone to including more spurious labels; for example, the label ``LS2: Blizzard Entertainment" (a software company), appears occasionally alongside ``LS2: Blizzard." These spurious labels were rare, and they were addressed with a high-pass filter on the labels' empirical document frequency $\docfreq_\termidx$, given by
\begin{align}
    \docfreq_\termidx := \frac{\countimages_\termidx}{\numimages}\label{eq:docfreq}
\end{align}
where $\countimages_\termidx$ is the number of images in the data set in which the label $\termidx$ appears, and $\numimages$ is the total number of images in the data set. The cutoff for the high-pass filter is set at $\docfreq_\termidx = 10^{-5}$, and was chosen heuristically. We considered that spurious labels may show up once or twice per camera; thus, we set the cutoff at a baseline average rate of three images per camera, $\sim0.01\%$, and consider only labels that appear in at least $0.01\%$ of the total images. In addition, we removed labels related to ``Massachusetts Department of Transportation," as those labels are likely due to the ``massDOT" watermark in the lower right corner, and not the actual scene content. We note that this data cleaning can likely be improved with a more careful and targeted approach, such as a term whitelist which contains only relevant labels of interest. However, for the coarse-grained analyses presented in this paper, we find that our described approach is sufficient in removing the majority of the spurious labels.

\section{Methods}
\label{sec:methods}
This section presents the methods used in our analysis. We formulate the \acf{BoLW} model in Sec.~\ref{sub:bolw}; introduce the \emph{per-camera idf} data weighting scheme in Sec.~\ref{sub:rescale}; construct the \emph{image-label} matrix representation in Sec.~\ref{sub:img_label}; and describe the \ac{LDA} topic model in Sec.~\ref{sub:topic_id}.

\subsection{Bag-of-Label-Words}
\label{sub:bolw}
Consider a vector space, $\labelset$, where each dimension corresponds to an individual label in the label vocabulary. The dimension of $\labelset$---the number of terms in the label vocabulary---is denoted $\numlabels$. Any weighted list of labels can be represented as a vector in this label vector space, denoted $\labelvec \in \labelset$, where the nonzero components of $\labelvec$ correspond to the weight of the respective label in the list. Let $\termidx$ index the vocabulary, and thus, let $\labelvec^\termidx$ represent the component of the $\labelvec$ corresponding to term $\termidx$. This vector representation is analogous to the \ac{BoW} vector space model of documents in \ac{NLP}, which represents documents as vectors where each component corresponds to the number of occurrences of a given word in the document \cite{Manning2009a}; hence, we refer to our model as \ac{BoLW}\footnote{There is a related \ac{BoW} model in computer vision, called Bag-of-Visual-Words (BoVW); however, BoVW uses pixel features as its ``words"}. 

In our model, an \emph{image}, $\image_\imageidx$, in a \emph{corpus}, $\imageset$, is represented as the tuple $\image_\imageidx = \left(\camera_\imageidx, \timestamp_\imageidx, \labelvec_\imageidx \right)$, where:
\begin{itemize}
    \item $\imageidx \in \{1, \dots, \numimages\}$ indexes the image
    \item $\camera_\imageidx$ denotes its originating camera
    \item $\timestamp_\imageidx$ denotes its timestamp
    \item $\labelvec_\imageidx$ is its \acl{BoLW} label vector
\end{itemize}
The \acl{BoLW} vector model is defined as follows:
\begin{itemize}
    \item A \emph{label word}, $\labelword_\termidx$, is defined as a single label in the label vocabulary, indexed by $\termidx \in \{1, \dots, \numlabels\}$. $\labelword_\termidx$ is a unit-basis vector in $\labelset$ whose $\termidx^\text{th}$ component equals one, and all other components equal zero.
    \item A \emph{bag of label words} associated with image $\imageidx$ is a vector $\labelvec_\imageidx \in \labelset$. Any weighted list of labels describing an image $\imageidx$ can be represented as a bag of label words by setting the respective weights of $\labelvec_\imageidx$ equal to the weights of the corresponding list elements.
    \item The total \emph{weight}, $\weight_\imageidx$, of bag $\labelvec_\imageidx$ is defined as its $L^1$-norm: $\weight_\imageidx := \Vert \labelvec_\imageidx \Vert_1 = \sum_\termidx |\labelvec_\imageidx^\termidx|$
\end{itemize}
While this paper focuses on the application of the \ac{BoLW} model for traffic camera image analysis, our approach can be generically applied to any corpus of discretely labeled images.

\subsection{Label Reweighting}
\label{sub:rescale}
We note that extremely common labels do not necessarily contribute much operationally useful information about the image contents. For example, labels such as ``Road" and ``Asphalt" appear extremely frequently in the Boston data set. While these labels are not incorrect---the images from freeway cameras do indeed contain roads made of asphalt---they are also not particularly informative, as we expect that most images from a traffic camera contain a road. Thus, we would like to attenuate the weight of labels which occur extremely frequently. We address this with the \ac{tf-idf} weighting scheme, described below, to rescale each image's label weights based on each label's rarity for each camera.

The \ac{tf-idf} weighting scheme is a heuristic used in \ac{NLP} to reweight terms in the \ac{BoW} vector to account for the natural difference in term prevalence in a language \cite{Manning2009a}. Terms that are commonly used in a language will highly represented in any given document, simply due to their prevalence in the language, regardless of their relevance to the subject matter of the document. These extremely common terms can end up dominating the weight of a bag, and thus, it may be desirable to attenuate them. The \ac{tf-idf} weight is computed as the product of its two titular components: the term frequency (tf) and the inverse document frequency (idf) \cite{Manning2009a}. The term frequency of a given document $\image$ and term $\termidx$ corresponds to the number of occurrences of the term within the document; in our case, our term frequency for image $\image$ and label $\termidx$ is given simply by the binary variable:
\begin{align}
    \text{tf}(\imageidx, \termidx) = \begin{cases}
    1 & \text{if image $\imageidx$ has label $\termidx$}\\
    0 & \text{otherwise}
    \end{cases}\label{eq:tf_def}
\end{align}
The inverse document frequency (idf) of a term $\termidx$ is typically computed as the negative logarithm the empirical document frequency: $ \text{idf}(\termidx) = -\log(\docfreq_\termidx) = \log\left(\frac{\numimages}{\countimages_\termidx}\right).$
We use a variant of idf, which we term the \emph{per-camera idf}, computed as:
\begin{align}
    \widetilde{\text{idf}}(\termidx, \camera) = \log\left(\frac{\numimages_\camera}{\countimages_\termidx}\right) \label{eq:percam_idf}
\end{align}
where $\numimages_\camera$ is the total number of images for camera $\camera$. The per-camera idf considers the relative rarity of a label $\termidx$ within the context of the other images from that camera. This is motivated by the fact that the label distributions are different across cameras; for example, the presence of the label ``Snow" is more unusual and notable for images from a camera in a tunnel than those from a camera out in the open.

\subsection{Image-label matrix}
\label{sub:img_label}
We construct the $\numimages \times \numlabels$ \emph{image-label} matrix, denoted $\labelmat$, by vertically concatenating row vectors $\labelvec_\imageidx$ for all images $\imageidx \in \{1\dots \numimages\}$, where the components of $\labelvec_\imageidx$ are the per-camera \ac{tf-idf} values, given by:
\begin{align}
    \labelvec_\imageidx^\termidx = \text{tf}(\imageidx, \termidx) \times \widetilde{\text{idf}}(\termidx, \camera_\imageidx).
\end{align}
Each row of $\labelmat$ corresponds to an image, and each column correspond to a label. This \emph{image-label} matrix is analogous to the \emph{document-term} matrix in \ac{NLP}, and in general, our usage of the terms ``image" and ``label" in this paper correspond to ``document" and ``term" respectively in the \ac{NLP} literature. 

The image-label matrix representation of the data set presents the data in a familiarly-structured form: a matrix of $\numimages$ observations of an $\numlabels$-dimensional system. Thus, the image labeling process transforms the unstructured traffic camera data into a conventional, structured $\numlabels$-dimensional time series analysis problem. However, as a label vocabulary can span hundreds to thousands of words, the dimension of $\labelset$ may still be prohibitively large for human interpretation and certain computations. Since $\labelmat$ is simply a $\numimages \times \numlabels$ matrix, many conventional matrix compression techniques can be applied to the data to reduce its dimensionality. However, depending on the technique, the compressed representation may not be semantically meaningful; this ends up forgoing much of the advantages in interpretability by representing the data as labels in the first place. Thus, we focus on topic models, which decompose the data into semantically distinct and interpretable topics; in particular, the \acl{LDA} topic model.

\subsection{Topic Identification via \acl{LDA}}
\label{sub:topic_id}

\begin{figure}[tb]
    \centering
    \includegraphics[width=0.5\linewidth]{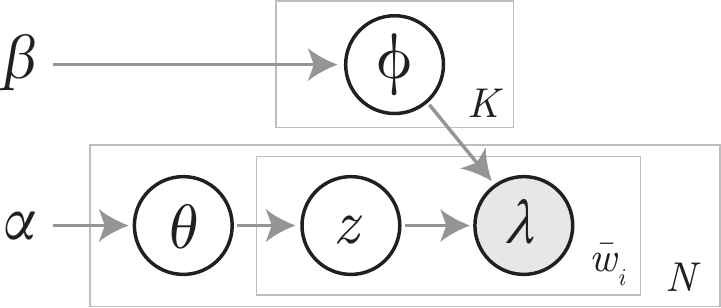}
    \caption{Graphical representation of the \ac{LDA} model structure. Each of the boxes (plates) represent a repeated component; the variable in the lower right hand corner of each plate indicates the number of copies. The outer plates represent each bag of label words in the corpus, and the inner plate represents each label word added to the bag. Grey-filled circles represent observed variables, whereas white-filled circles represent latent variables.}
    \label{fig:lda_plate}
\end{figure}
\begin{algorithm}[h!]
\SetKwInOut{Input}{input}
\SetKwInOut{Output}{output}
\Input{Target weight, $\overline{\weight}_\imageidx$, of bag $\labelvec_\imageidx$\newline
image-topic prior hyperparamater $\doctopicparam$ \newline
topic-label prior hyperparamater $\topicwordparam$}
\Output{bag of label words $\labelvec_\imageidx$}
    Initialize $\labelvec_\imageidx=\mathbf{0}$\\
    Draw $\doctopicdist \sim \text{Dirichlet}(\doctopicparam)$\\
    Draw $\topicworddist \sim \text{Dirichlet}(\topicwordparam)$\\
    \While{$\weight_\imageidx < \overline{\weight}_\imageidx$}{
    Draw topic $\topic \sim \text{Multinomial}(\doctopicdist^\imageidx)$\\
    Draw a label word $\labelword \sim \text{Multinomial}(\topicworddist^\topic)$\\
    Add $\labelword$ to the bag: $\labelvec_\imageidx = \labelvec_\imageidx + \labelword$
    }
\caption{\ac{LDA} \ac{BoLW} generation procedure}
\label{alg:lda}
\end{algorithm}

\acf{LDA} is a hierarchical Bayesian topic model for document generation in \ac{NLP}, first posed by \citeauthor{Blei2003a} \cite{Blei2003a}. \ac{LDA} represents documents as random mixtures of topics, denoted $\doctopicdist$, where each topic is, in turn, a probability distribution over words, denoted $\topicworddist$. \citeauthor{Griffiths2004} present a variant that includes an additional Dirichlet prior over on the topic-word distribution $\topicworddist$ \cite{Griffiths2004}. We adapt this variant of \ac{LDA} to \acp{BoLW} below, and visualize it in plate notation in Figure~\ref{fig:lda_plate}.
Each bag of label words $\labelvec_\imageidx$ in a corpus $\imageset$ is generated by the \ac{LDA} model with the procedure described in Algorithm~\ref{alg:lda}.
The target weights $\overline{\weight}_\imageidx$ are set exogenously based on the empirical bag weights in the corpus; in addition, the target weights are rounded to the nearest integer, since the model requires an integer number of copies of the innermost plate in Fig.~\ref{fig:lda_plate}.
The \emph{image-topic} distribution is denoted $\doctopicdist$ and characterized by the $\numtopics-$dimensional hyperparameter $\doctopicparam$, and $\doctopicdist^\imageidx = \doctopicdist(\topic|\imageidx)$. A topic is denoted $\topic \in \{1,\dots,\numtopics\}$, where $\numtopics$ is the total number of topics, set exogenously. The \emph{topic-label} distribution is denoted $\topicworddist$ and characterized by the $\numlabels$-dimensional hyperparameter $\topicwordparam$. The distribution of labels for a given topic is denoted $\topicworddist^\topic = \topicworddist(\labelword|\topic)$.

\begin{figure*}[htb]
    \centering
    \subfloat[Sample of \ac{LDA} topics, and their respective highest probability labels in descending order]{\footnotesize
    \begin{tabular}{|l||p{1in}||l||p{1in}||l|}
    \hline
\thead[l]{\textbf{Topic 1}: Wintry\\ Conditions}     & \textbf{Topic 3}: Night             & \textbf{Topic 4}: Car           & \textbf{Topic 8}: Intersection              & \textbf{Topic 10}: Error     \\ \hline
 LS1: snow                   &  LS1: night              &  LS2: Car            &  LS1: thoroughfare               &  LS1: white       \\
 LS2: Snow                   &  LS2: Street             &  LS1: car            &  LS2: Intersection               &  LS1: material    \\
 LS2: Phenomenon             &  LS2: Night              &  LS1: vehicle        &  LS1: intersection               &  LS1: technology  \\
 LS1: phenomenon             &  LS1: street light       &  LS2: Vehicle        &  LS2: Asphalt                    &  LS2: Webcam      \\
 \makecell[l]{LS1: geological\\\quad\quad phenomenon}  &  \makecell[l]{LS2: Mode of\\\quad\quad transport}  &  \makecell[l]{LS1: motor\\\quad\quad vehicle}  &  \makecell[l]{LS2: Controlled-access \\\quad\quad highway}  & \makecell[l]{LS1: circle}\\\hline
\end{tabular} \label{tab:selected_topics}} \hspace{2em}
\subfloat[Error message that is shown when a live feed for a camera is unavailable]{
    \raisebox{-0.25in}{\includegraphics[width=.2\linewidth]{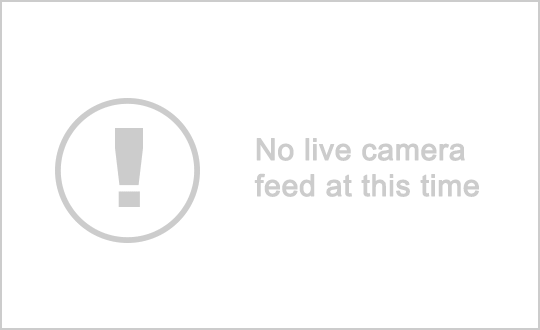}}\label{fig:no_feed}}
    \caption{Selected \ac{LDA} topics \protect\subref{tab:selected_topics} and unavailable feed error message \protect\subref{fig:no_feed}}
\end{figure*}

Given the hyperparameters and a corpus of data $\imageset$, we would like to infer the most likely \emph{image-topic} distribution, $\doctopicdist$ and \emph{topic-word} distribution, $\topicworddist$, which maximizes the posterior probability $p(\doctopicdist, \topicworddist | \imageset, \doctopicparam, \topicwordparam)$. We achieve this with the online variational Bayes algorithm presented in \cite{Hoffman}. We assume symmetric priors on $\doctopicdist$ and $\topicworddist$ with constant hyperparameter values $\doctopicparam=\frac{50}{\numtopics}$ and $\topicwordparam=0.1$ based on \cite{Griffiths2004}. For a detailed treatment of \ac{LDA} and other probabilistic topic models, we recommend \cite{Steyvers2010}.

\section{Results}
\label{sec:results}

\subsection{Topics}
\label{sub:topics}
In this section, we discuss the \ac{LDA} topics, and present the top labels for a selection of representative topics. For the \ac{LDA} decomposition in this example, we chose $\numtopics$ qualitatively: we wanted the number of topics to be small enough to be easily visualized and interpreted, but large enough to distinguish between the following semantic categories in the data: i.) snow storms; ii.) the day/night cycle; iii.) traffic; iv.) physical infrastructure; and v.) error messages. For illustrative purposes, we found that $\numtopics=10$ provides a manageable number of semantically distinct and relevant topics. In practice, the selection of the optimal number of topics, $\numtopics$, depends on the target application; in \cite{Griffiths2004}, \citeauthor{Griffiths2004} propose selecting $\numtopics$ which maximizes the log-likelihood of the data.

We present in Table~\ref{tab:selected_topics} five representative topics---one for each of the aforementioned categories---along with their respective highest-probability labels. We have followed common practice and manually named topics---\emph{a posteriori}---based on domain-specific knowledge of the data, for ease of reference. Note that Topic 10: ``Error" is a special edge case; the topic appears only on the image depicted in Figure~\ref{fig:no_feed}, which is returned when the web service is unable to provide the current feed for a given camera. This allows us to conveniently identify error images in the data; this is notable, as the \ac{LDA} is not explicitly trained to identify error images. This suggests that the \ac{LDA} decomposition may be a suitable transformation for image classification tasks, which we will explore in future work.


\subsection{Topic time-series}
\label{sub:trends}
The \ac{LDA} image-topic distributions $\doctopicdist^\imageidx$ can be interpreted as projections of the \ac{BoLW} vector, $\labelvec_\imageidx$, onto the \ac{LDA} topic space. By plotting the probability of each topic for an image $\doctopicdist^\imageidx(\topic)$ at its timestamp $\timestamp_\imageidx$, we can construct a set of time-series for each topic and camera. Figures in this section were plotted with the data averaged into 15-minute bins. We examine the detection of weather events in Section~\ref{sub:weather_events}; and the effect of infrequent events on weekly traffic patterns in Section~\ref{sub:weekly_patterns}.

\subsubsection{Detecting weather events}
\label{sub:weather_events}
We identified notable weather events during the data collection period via the \ac{NOAA-GHCN} data set \cite{menne2012global}. We define a notable weather event as one with at least 1" of snow or rainfall. We identified six such events during the collection period: two with snow and rain on 2017-12-25 and 2018-01-04 (the ``bomb cyclone"); two with only rain on 2018-01-13 and 2018-01-23; and two with only snow on 2018-01-17 and 2018-01-30.
\begin{figure}[h!]
    \centering
    \subfloat[Time series of ``LS1: snow" label]{\includegraphics[width=0.8\linewidth]{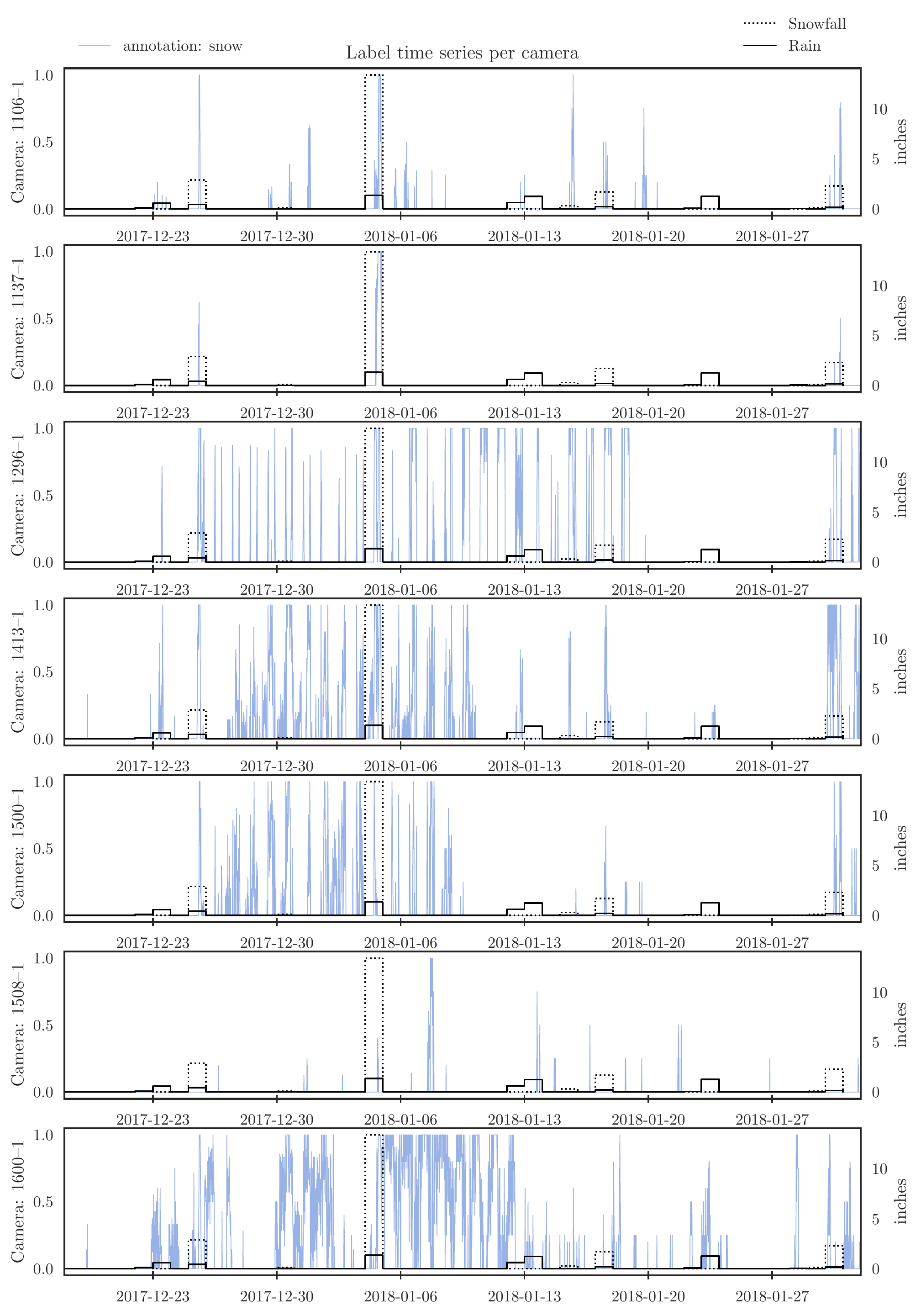}\label{fig:snow_LS1_ts}}
    \hspace{1em}
    \subfloat[Time series of Topic 1: ``Wintry conditions" ]{\includegraphics[width=0.8\linewidth]{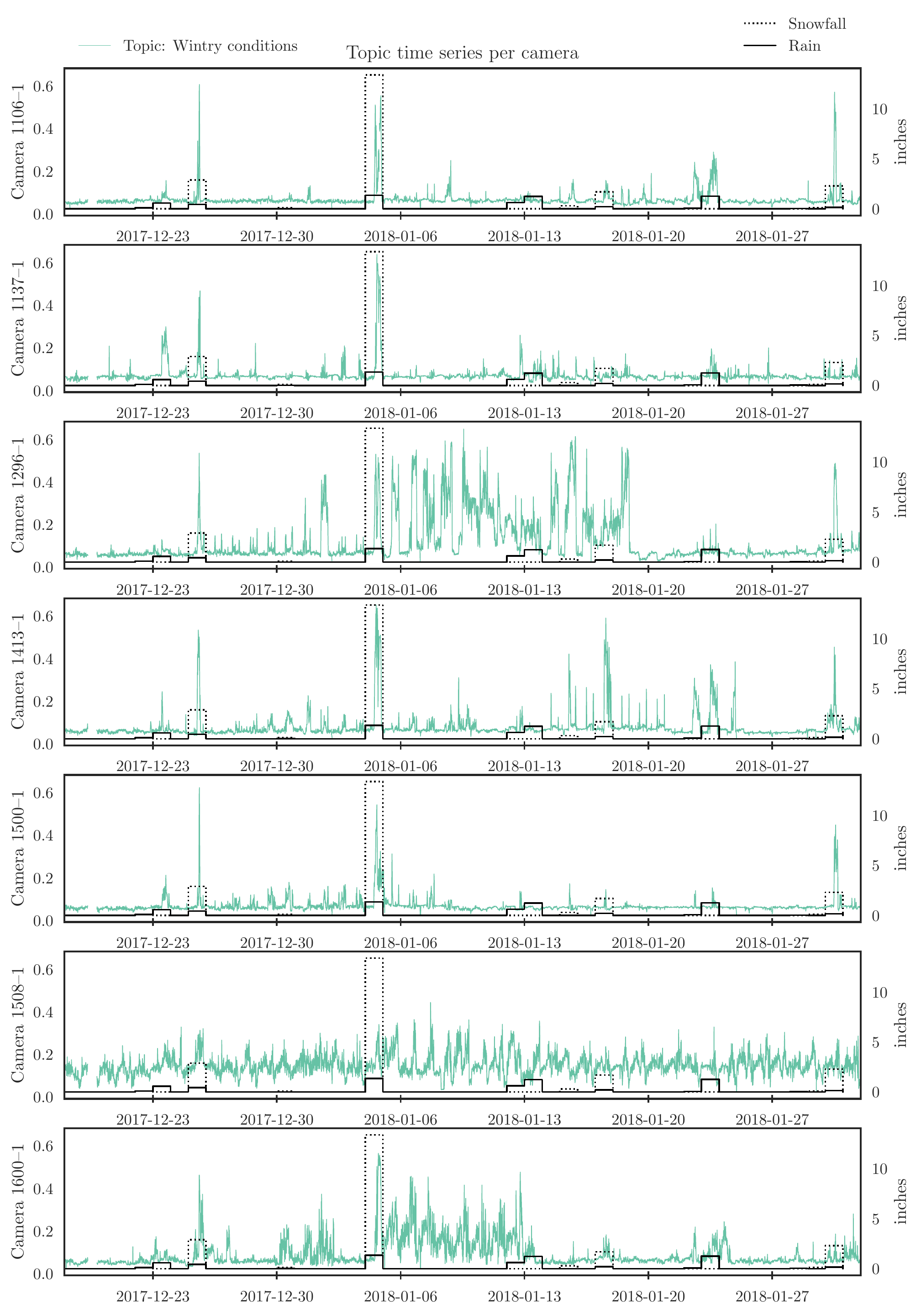}\label{fig:snow_topic_ts}}
    \caption{Comparison of the sensitivity of the ``LS1: snow" label \protect\subref{fig:snow_LS1_ts} and Topic 1: ``Wintry conditions" \protect\subref{fig:snow_topic_ts} to winter storms.}
    \label{fig:snow_timeseries}
\end{figure}

We first consider the simple approach of using a single label: ``LS1: snow" to detect weather events. Its time series---the column of $\labelmat$ corresponding to ``LS1: snow"---is plotted for each camera in Figure~\ref{fig:snow_LS1_ts}. For most cameras, the ``LS1: snow" time series is sensitive to the weather events that include snow. The exception is camera 1508--1; however, this is to be expected, as this camera is located in a underpass that is isolated from the elements (see Fig.\ref{fig:cameras}). We note two issues: first, the label is not sensitive to weather events that are exclusively rain (only two cameras, 1413--1 and 1600--1, have any signal during the 2018-01-23 rain event). This, however, can be overcome by manually including additional labels, such as the ``rain" LS1 and LS2 labels. Second, there are significant daily recurrences of the ``LS1: snow" label in many cameras following the weather events of 2017-12-25 and 2018-01-04. The recurrence corresponds to the detection of accumulated snow on the ground. The recurrence can make it difficult to differentiate new snowfall from recurring detection of existing snow (cf. cameras 1296--1, 1413--1, 1500--1). In addition, while the snowfall during the ``bomb cyclone" event was much greater than the 2017-12-25 event, this is not directly apparent in the label time series. 

We now demonstrate how \ac{LDA} ameliorates both of the above issues. Figure~\ref{fig:snow_topic_ts} illustrates the time series for Topic 1: ``Wintry conditions" for each camera. These time series are still sensitive to the snowfall events (except camera 1508--1, as expected), but unlike the ``LS1: snow" time series, they also show small signals in response to the rainfall events. In addition, we observe that the effect of the daily detection of accumulated snow is significantly attenuated compared to the signal of the ``LS1: snow" time series. Cameras 1296--1 and 1600--1 still exhibit the daily recurrence trait---albeit to a lesser degree---particularly after the large ``bomb cyclone" blizzard. These two cameras are angled closer to the road than the other outdoor cameras, and thus they see more of the ground. As such, we expect the effect of accumulated snow to be more pronounced for these two cameras. Additionally, whereas the ``LS1: snow" time series did not clearly show any difference in magnitude in the label recurrence between the 2017-12-25 and 2018-01-04 events; the magnitude of the ``Wintry conditions" topic recurrence following the ``bomb cyclone" 2018-01-04 storm is larger, consistent with the storm's greater snowfall.


\subsubsection{Weekly traffic patterns and disturbances}
\label{sub:weekly_patterns}
In this section, we analyze the weekly daytime traffic patterns using the time series of Topic 4: ``Car" from camera 1508--1. This camera is selected due to its location in an underpass, which isolates it from the direct effects of weather; in addition, we did not observe any camera angle changes during the observation period. Its isolation from weather and static camera angle ensures that any changes in the traffic pattern are due to changes in demand, and not a side effect of reduced visibility or change in camera angles.

\begin{figure}[htb]
    \centering
    \subfloat[Week of Christmas: Monday, December 25$^\text{th}$, 2017 ]{\includegraphics[width=\linewidth]{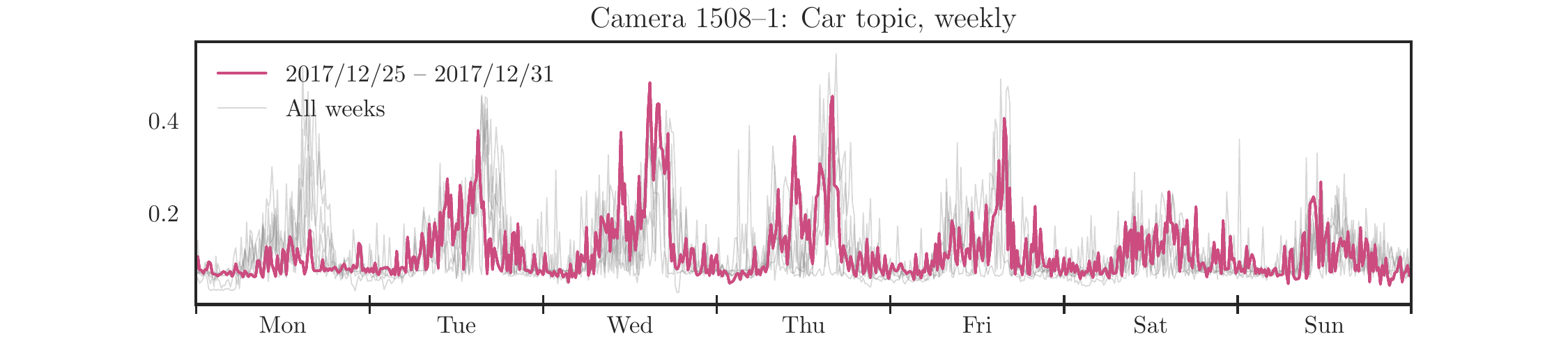}\label{fig:wk_ts_christmas}}
    
    \subfloat[Week of Bomb Cyclone: Thursday, January 4$^\text{th}$, 2018]{\includegraphics[width=\linewidth]{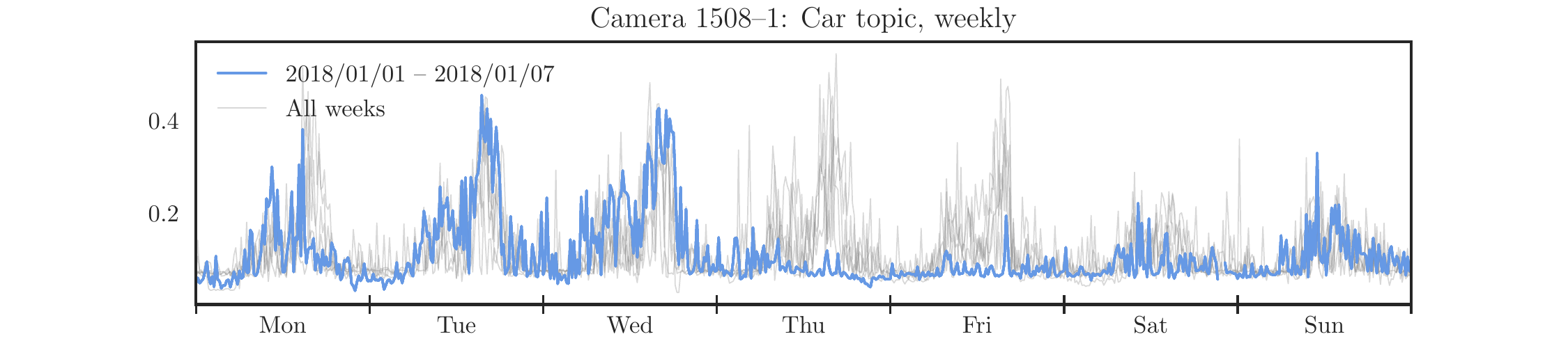}\label{fig:wk_ts_bomb_cyclone}}
    \caption{The ``Car" topic of camera 1508--1 captures the weekday-weekend traffic demand cycle. The time series signal shows lower readings for the days corresponding to the Christmas holiday (\ref{fig:wk_ts_christmas}) and the ``Bomb Cyclone" winter storm (\ref{fig:wk_ts_bomb_cyclone}).}
    \label{fig:wk_ts_1508}
\end{figure}

We illustrate in Figure~\ref{fig:wk_ts_1508} the weekly time series of the ``Car" topic for camera 1508--1. The light grey lines represent the data for all weeks, whereas the highlighted lines in \ref{fig:wk_ts_christmas} and \ref{fig:wk_ts_bomb_cyclone} highlight the weeks of the Christmas holiday, and the ``bomb cyclone" storm respectively. We observe the weekday-weekend traffic pattern of greater demand during the weekdays than on the weekends. Additionally, we observe a larger peak during the evening rush hours on weekdays. This is consistent with the camera's location on an on-ramp to I-93 South, leading out of Boston (Fig. \ref{fig:cameras}). Indeed, though the labels do not explicitly encode the car counts, we demonstrate that some magnitude information can be inferred binarized label data.

In Figure~\ref{fig:wk_ts_christmas}, we observe that on Christmas, which occurred on a Monday, the ``Car" topic was significantly lower than on other weeks. The rest of the week, however, was not significantly different than average. This is consistent with a reduction in traffic on Christmas day, as it is a national holiday, and many institutions and businesses are closed. We also note that while New Year's day (Monday,  Figure~\ref{fig:wk_ts_bomb_cyclone}) is also a national holiday, we observe more traffic on New Year's than on Christmas. This is likely because in the US, many businesses are closed on Christmas but are open on New Year's, albeit with limited hours.

We observe the effect of the bomb cyclone in Figure~\ref{fig:wk_ts_bomb_cyclone}. Whereas the snow did not start falling until the evening of January 4$^\text{th}$, the observed counts are at near zero starting in the morning, and remain there until Friday evening. Even then, Friday evening and Saturday have lower readings than usual; only on Sunday do things return to normal. The low ``Car" topic weights correspond with the City's declaration of Snow Emergency and Parking Ban, which was in effect between 7~a.m., January 4$\text{th}$--5~p.m., January 5$^\text{th}$ \cite{Andersen}.


\section{Conclusion}
\label{sec:discussion}
In summary, we presented: the \acl{BoLW} vector model for representing images in a semantic label space; the application of the \ac{LDA} topic model as a dimensionality reduction tool; and an analysis of freeway traffic cameras in the Boston metropolitan area using these techniques. We are able to distinguish between new snowfall and accumulated snow, and also to capture relative changes in magnitude in the example data set, despite using only binarized label data, which has potential application in data compression and privacy contexts.

We observe that disruptions, such as storms, manifest clearly in the topic time series. This offers a simple approach to performing change and anomaly detection on image data. Whereas most image change detection algorithms operate in the high-dimensional pixel space \cite{Radke2005}, our approach uses \acp{CNN} to transform the problem into the semantic space, where the \ac{BoLW} representations and \ac{LDA} dimension reductions allow for conventional univariate and multivariate algorithms to analyze the changes. Indeed, as traffic is an inherently dynamic phenomenon, pixels will always be changing due to passing vehicles. By representing the images in the semantic space, we can analyze phenomena that are difficult to parameterize in pixel space. We are currently working on automatic detection of anomalous traffic patterns using these techniques. 

Finally, we highlight the general applicability of our approach; we use the same methodology to sense both weather events and traffic patterns. This is a step toward to realizing the potential of general purpose sensing with cameras. While our constraint of using only binarized label data was intentionally restrictive to study the information content of labels on their own, we are exploring the use of traffic camera label data in conjunction with other detectors, such as loop detectors, in sensor fusion applications. 

\bibliographystyle{IEEEtranN}
\bibliography{references}



%



\end{document}